\RequirePackage{amsthm}
\documentclass[sn-apa]{sn-jnl}

\usepackage{graphicx}%
\usepackage{multirow}%
\usepackage{amsmath,amssymb,amsfonts}%
\usepackage{amsthm}%
\usepackage{mathrsfs}%
\usepackage[title]{appendix}%
\usepackage[dvipsnames]{xcolor}%
\usepackage{textcomp}%
\usepackage{manyfoot}%
\usepackage{booktabs}%
\usepackage{algorithm}%
\usepackage{algorithmicx}%
\usepackage{algpseudocode}%
\usepackage{listings}%
\usepackage{anyfontsize}%
\usepackage{natbib}%
\usepackage{nicefrac}%
\usepackage{xcolor}
\usepackage{placeins} 

\hypersetup{
    colorlinks=true,
    citecolor=NavyBlue,
    linkcolor=NavyBlue
}

\newcommand{\blue}{}
\newcommand{\bluetext}{}

\def\argmin{\operatornamewithlimits{arg\,min}}

\newcommand{\defeq}{\doteq}

\newcommand{\gramnplus}{K_{0}}

\newcommand{\newpart}{}

\newcommand{\BX}{\mathbb{X}}

\newcommand{\BI}{\mathbb{I}}

\newcommand{\CO}{\mathcal{O}}

\newcommand{\CH}{\mathcal{H}}

\newcommand*{\rom}[1]{\expandafter\@slowromancap\romannumeral #1@}
\def \CG{\mathcal{G}}

\def \CD{\mathcal{D}}

\newcommand{\tr}{^\mathrm{T}}  
\newcommand{\RR}{\mathbb{R}}

\newcommand{\norm}[1]{\left\lVert#1\right\rVert}

\def\argmin{\operatornamewithlimits{arg\,min}}        

\theoremstyle{thmstyleone}%
\newtheorem{theorem}{Theorem}

\theoremstyle{thmstyletwo}%

\theoremstyle{thmstylethree}%
\newtheorem{definition}{Definition}%
\newtheorem{assumption}{\bf A\!\!}
\newtheorem{assumption_b}{\bf B\!\!}
\newtheorem{lemma}{Lemma}

\makeatletter
\newcommand\fs@spaceruled{\def\@fs@cfont{\bfseries}\let\@fs@capt\floatc@ruled
  \def\@fs@pre{\vspace{0.5\baselineskip}\hrule height.8pt depth0pt \kern3pt}%
  \def\@fs@post{\kern3pt\hrule\vspace{-2mm}\relax}%
  \def\@fs@mid{\kern3pt\hrule\kern2pt}%
  \let\@fs@iftopcapt\iftrue}
\makeatother

\begin{document}

\title[Article Title]{Single Image Inpainting and Super-Resolution with Simultaneous Uncertainty Guarantees by 
Universal 
Reproducing 
Kernels}

\author[1,2]{\fnm{B\'alint} \sur{Horv\'ath}}\email{\color{NavyBlue}balint.horvath@sztaki.hu}
\author*[1,3]{\fnm{Bal{\'a}zs Csan{\'a}d} \sur{Cs{\'a}ji}}\email{\color{NavyBlue}balazs.csaji@sztaki.hu}

\affil*[1]{Institute for Computer Science and Control (SZTAKI), Hungarian Research Network (HUN-REN), Budapest, Hungary, 1111}
\affil*[2]{Institute of Mathematics, Budapest University of Technology and Economics (BME), Budapest, Hungary, 1111}
\affil*[3]{Dept.\! of Probability Theory and Statistics, Institute of Mathematics, Eötvös Loránd University (ELTE), Budapest, Hungary, 1117}

\abstract{The paper proposes a statistical learning approach to the problem of estimating missing pixels of images, crucial for image inpainting and super-resolution problems. One of the main novelties of the method is that it also provides uncertainty quantifications together with the estimated values. Our core assumption is that the underlying data-generating function comes from a Reproducing Kernel Hilbert Space (RKHS). A special emphasis is put on band-limited functions, central to signal processing, which form Paley-Wiener type RKHSs. The proposed method, which we call Simultaneously Guaranteed Kernel Interpolation (SGKI), is an extension and refinement of a recently developed kernel method. An advantage of SGKI is that it not only estimates the missing pixels, but also builds non-asymptotic confidence bands for the unobserved values, which are simultaneously guaranteed for all missing pixels. We also show how to compute these bands efficiently using Schur complements, we discuss a generalization to vector-valued functions, and we present a series of numerical experiments on various datasets containing synthetically generated and benchmark images, as well.}


\maketitle

\section{Introduction}\label{sec1}
The paper proposes a {\em statistical learning} method for two fundamental {\em image processing} problems, while also providing 
{\em uncertainty quantification} for the solutions.

The problem of {\em image restoration} is to create a higher quality image from a lower quality measurement. Examples for the lack of quality include missing, damaged pixels due to noise, blur, or compression \citep{petrou2010image,8964437}. If the method is based solely on one image, it is called {\em single image restoration}; and when the objective is to estimate the values of missing pixels, it is called {\em single image inpainting} \citep{elharrouss2020image}.
There are several methods for these problems, from total variation based techniques to methods using biharmonic equations \citep{acharya2005image}, and recently {\em deep learning} approaches also became popular \citep{8627954,9454311}. 

A closely related task is the so-called {\em single image super-resolution}, which aims to create a higher resolution image from a lower resolution one \citep{milanfar2017super}. A wide range of super-resolution methods have been developed over the years, from bicubic interpolation to Lanczos resampling, but recent approaches implementing {\em deep learning}, such as generative adversarial network (GAN) or convolutional network based methods, produce the state-of-the-art results \citep{ledig2017photo,li2019feedback}. On the other hand, most of the available methods do not come with strict theoretical and uncertainty guarantees. They generally do not provide information about how accurate their pixel estimates are, they do not quantify the uncertainty of their outcomes.

{\bluetext There are many application domains, from autonomous driving, cosmology, and geospatial imaging, to 
art restoration, forensic analysis and medical imaging \citep{huang2024review}, for which quantifying the uncertainty of the reconstructed pixels is crucial. For example, in medical diagnoses, such as reconstructing missing pats of MRI scans or removing artifacts from X-ray images, having point estimates for the missing pixels is not sufficient. In these cases, just providing a realistic-looking restoration could even be misleading, potentially leading to misdiagnosis. Clinicians need to know how confident the inpainting algorithm is about its reconstruction to assess where the solutions may be unreliable. Showing confidence intervals for pixels can help to determine the need of further tests or alternative diagnosis types. It is also known that changing few pixel values can have radical effects on deep learning models \citep{szegedy2013intriguing}, and even one-pixel modifications can lead to class label changes \citep{su2019one}. These issues further underscore the need to quantify the uncertainty of the estimates.}

The main contribution of this paper is that it proposes a statistical learning approach, called {\em Simultaneously Guaranteed Kernel Interpolation} (SGKI), to single image inpainting and super-resolution problems that not only provides estimates for the unknown pixel values, but it also constructs {\em non-asymptotic}, {\em non-parametric} {\em confidence regions} which are {\em simultaneously guaranteed} for all missing pixels. It is based on the theory of {\em Reproducing Kernel Hilbert Spaces} (RKHSs), see \citep{paulsen2016introduction}, with a special emphasis on {\em Paley-Wiener} (PW) spaces \citep{avron2016quasi}. The importance of PW spaces for these types of image processing problems can be explained by the fact that they contain band-limited functions which are central to the theory of signal processing. We build on the results of \citep{csaji2022nonparametric,csaji2023improving} which exploit the properties of {\em minimum norm interpolants} in RKHSs. A variant of this idea, where a deterministic setting was used with a priori known bounds on the observation noises and on the norm of the data-generating function, was also studied in \citep{scharnhorst2022robust}. 

The structure of the paper is as follows. First, in Section \ref{sec:rkhs} a brief introduction to RKHSs and Paley-Wiener spaces are given. Section \ref{sec:problem} formalizes the problem setting and our main objectives. In Section \ref{sec:construction}, we first provide the construction for the case, when a (high probability) upper bound is available for the kernel norm of the underlying data-generating function. 
A possible extension to color images, involving vector-valued functions, is also discussed. Then, in Section \ref{sec:kernelnorm} we study how to construct an upper bound of the kernel norm from the image itself. Section \ref{sec:schur} addresses the problem of reducing the computational complexity of the method by recursively computing the inverse of the kernel matrix. Finally, numerical experiments with comparisons are shown in Section \ref{sec:experiments}, while Section \ref{sec:conlusions} summarizes and concludes the paper.

\section{Paley-Wiener Spaces and Universal Kernels}
\label{sec:rkhs}
Kernel methods are widely used in machine learning, signal processing, statistics and related fields \citep{berlinet2004reproducing,paulsen2016introduction}. Here we provide a brief summary of the most important concepts used in the paper.
\vspace{-2mm}
\begin{definition}[Reproducing Kernel Hilbert Spaces]
\textit{Let $\mathbb{X} \neq \emptyset$ be an arbitrary set. A Hilbert space $\CH$ of functions $f: \mathbb{X} \to \mathbb{R}$, with an inner product $\langle\cdot,\cdot\rangle_{\CH}$, is a Reproducing Kernel Hilbert Space (RKHS), if each Dirac functional,
$\delta_z: f \to f(z)$, is 
continuous, for all $z \in \BX$.}
\end{definition}
\vspace{-2mm}

For every RKHS, a unique {\em kernel} map $k: \mathbb{X} \times \mathbb{X} \to \mathbb{R}$ can be constructed, which has the \textit{reproducing property}, $\langle k(\cdot,z),f \rangle_\mathcal{H} = f(z),$ for each $z \in \mathbb{X}$ and $f \in \CH$. Kernels are always {\em symmetric} and {\em positive definite} functions. 
The Moore-Aronszajn theorem guarantees that conversely, for every 
symmetric and positive definite function, a unique RKHS exists for which it is its reproducing kernel \citep{paulsen2016introduction}.

Typical kernels include, for example, the Gaussian kernel defined by $k(x_1, x_2) = \exp(\nicefrac{-\|x_1-x_2\|^2}{2 \sigma^2})$, with $\sigma > 0$, the 
polynomial kernel, $k(x_1, x_2) = (\left< x_1, x_1 \right> + c)^p$, with $c \geq 0$ and $p \in \mathbb{N}$, and the sigmoidal kernel, 
$k(x_1, x_2) = \tanh(a \left< x_1, x_2 \right> + b)$ for some $a, b \geq 0$, where $\left< \cdot, \cdot \right>$ denotes the standard Euclidean inner product.

The {\em Gram} matrix of kernel $k$ w.r.t.\ inputs $x_1, \dots, x_n \in \mathbb{X}$ is 
$K_{i,j} \defeq k(x_i,x_j)$, for all {\newpart  $i, j \in [n] \doteq \{1,\dots, n\}$}.
Matrix $K$ is always positive semi-definite. A kernel is called {\em strictly} positive definite, if its Gram matrix is positive definite for all {\em distinct} $\{x_k\}$.

A very important example of RKHSs are formed by band-limited functions. As this type of functions play a fundamental role in the theory of signal processing (see, for example, the Nyquist–Shannon sampling theorem), these spaces also have a special importance for our method. In fact, this choice of spaces will allow us to estimate how ``smooth'' the underlying function is (which is measured by the kernel norm).

\vspace{-2mm}
\begin{definition}[Paley-Wiener Spaces]
\textit{A Paley-Wiener space $\mathcal{H}$ is a subspace of $\mathcal{L}^2 (\mathbb{R}^d)$, i.e., the space of square-integrable functions on $\mathbb{R}^d$, where for each $\varphi \in \mathcal{H}$ the support of the Fourier transform of $\varphi$ is included in a given hypercube $[-\eta, \eta\hspace{0.3mm}]^d$, where $\eta > 0$ is a hyper-parameter.}
\end{definition}
\vspace{-2mm}

Paley-Wiener spaces are
RKHSs \citep{berlinet2004reproducing} with the following {\em strictly} positive-definite {\em reproducing kernel} function. For all $u, v \in \mathbb{R}^d,$ we have
\vspace{-1mm}
$$
k(u,v) \,\doteq \,  \frac{1}{\pi^{d}} \prod_{j=1}^d \frac{\sin(\eta(u_j-v_j))}{u_j-v_j},
\vspace{2mm}
$$
where, for convenience, $\sin(\eta \cdot 0)/0$ is defined to be $\eta$. Note that as a Paley-Wiener (PW) space is a subspace of $\mathcal{L}^2$, it inherits its inner product and its norm.

An important question when choosing kernels is that what kind of functions can be represented or approximated by linear combinations of the chosen kernel. 

\vspace{-2mm}
\begin{definition}[Universal Kernels]
\textit{Let $\mathbb{X}$ be a metric space, let $\mathcal{Z} \subseteq \mathbb{X}$ be a compact subset and let $k:\mathbb{X} \times \mathbb{X} \to \mathbb{R}$ be a kernel, i.e., symmetric and positive definite. Let $\mathcal{K}(\mathcal{Z}) \doteq \overline{\text{span}}\{\,k(z,.):z\in \mathcal{Z}\,\}$, i.e., 
{\blue the closure w.r.t.\ the supremum norm of the linear span of all $k(z,.)$ functions, where $z\in \mathcal{Z}$}.
Let $\mathcal{C}(\mathcal{Z})$ be the set of all continuous $f:\mathcal{Z}\to\mathbb{R}$ type function. Kernel $k$ is called universal if and only if for all compact $\mathcal{Z} \subseteq \mathbb{X}$, we have $\mathcal{K}(\mathcal{Z}) = \mathcal{C}(\mathcal{Z})$.}
\end{definition}
\vspace{-2mm}

In other words, linear combinations of universal kernels can approximate  arbitrary well (in the supremum norm) continuous functions on any compact set, that is, they have the {\em universal approximating property}  \citep{micchelli2006universal}. 

A kernel $k: \mathbb{X} \times \mathbb{X} \to \mathbb{R}$ is called {\em translation invariant}, if there is a function $\psi:  \mathbb{X} \to \mathbb{R}$, such that $\forall\, x,y \in \mathbb{X}: k(x,y) = \psi(x-y).$ Let $\mathbb{X}$ be a metric space (this allows us to talk about the continuity and integrability of the kernels, as we can use the Borel $\sigma$-algebra induced by the metric of the space).  It is known that bounded, continuous, integrable, translation invariant, strictly positive definite kernels are universal \citep{sriperumbudur2011universality}. Based on this, both the Gaussian and the Paley-Wiener kernel are universal. 
It will also be important for us that continuous, universal kernels are strictly positive definite \citep{sriperumbudur2011universality}, consequently, their Gram matrices are invertible for distinct inputs.

\section{Problem Setting}
\label{sec:problem}

We treat images as functions having $d=2$ dimensional input vectors. The pixels are outputs of these functions at some observed inputs, typically distributed on a {\em grid}, for example, corresponding to the array of photosites of a digital camera. In this paper, for simplicity, we assume that the inputs come from $\CD \defeq [\hspace{0.5mm}0,1\hspace{0.3mm}] \times [\hspace{0.5mm}0,1\hspace{0.3mm}]$, and first we also assume that the outputs are scalars, hence we have {\em grayscale} images. More precisely, the pixel intensities are {\em centered} and {\em scaled}, therefore, they are from $[-1, 1\hspace{0.3mm}]$. 

Later, we will discuss how to extend the approach to multi-dimensional outputs, for example, to the case when the pixel values are from 
$[-1, 1\hspace{0.3mm}]^m$ describing RGB (red, green, blue) or CMYK (cyan, magenta, yellow, key / black) color codes.

Our fundamental assumptions on the available data can be summarized as
\vspace{-2mm}
\begin{assumption}
\label{A0} 
{\it We have a finite sample,
$(x_1, y_1), \dots, (x_n, y_n)$,
where $x_k \in \CD \defeq [\hspace{0.5mm}0, 1\hspace{0.3mm}]^d,$
$y_k \in [-1, 1\hspace{0.3mm}]$, and, $y_k \, = \, f_*(x_k),$
for $k \in [n] \doteq \{1, \dots, n\}$. 
We assume that the inputs $\{x_k\}$ are distinct, formally, $\forall\, i,j \in [n]:i \neq j \Rightarrow x_i \neq x_j$.
We call $f_*$ the ``true'' data generating function and 
assume that it belongs to a known RKHS $\CH$ with a continuous and universal kernel $k$.}
\end{assumption}
\vspace{-2mm}
We will gradually introduce more assumptions as needed.
Note that in the current version of the method we assume that the underlying ``true'' data generation function can be observed perfectly at the sample inputs. Therefore, we disregard observation noises including output quantization. We do so to simplify the presentation, but the approach can be extended to the case of noisy observations, as well, following the ideas in \citep{csaji2022nonparametric, csaji2023eysm, scharnhorst2022robust}.

Our two main {\em goals} are (1) to estimate the {\em missing pixel value} of any given (out of sample) query input point, as well as (2) to provide {\em uncertainty bounds} for our estimates. Albeit, there are several image processing methods to solve (1), but most of them do not come with uncertainty guarantees: they do not address (2). 

Formally, the goals are to construct a {\em point estimate} $\bar{f}$ of $f_*$ together with a {\em confidence band}, i.e., a function $I: \CD \to \mathbb{R}^2$, such that $I(x) = (\hspace{0.3mm}I_1(x), I_2(x)\hspace{0.3mm})$ specifies the {\em endpoints} of an interval estimate for $f_*(x)$ and contains $\bar{f}(x)$, for any $x \in \CD$. 
Hence, based on our dataset $\{(x_i, y_i)\}_{i=1}^{n}$, we need to construct function $\bar{f}$ and band $I$ with
\vspace{-1mm}
\begin{align*}
\forall\hspace{0.3mm} x \in \CD\;&:\;\;\, \bar{f}(x) \in  [\hspace{0.3mm}I_1(x), I_2(x)\hspace{0.3mm}], \\[1mm]
\nu(I)\;\,&\defeq \; \mathbb{P} \big(\, \forall\hspace{0.3mm} x \in \CD: {\newpart I_1(x) \leq f_*(x) \leq I_2(x)} \,\big) \, \geq \, 1- \gamma,\\[-6.5mm]
\end{align*}
where $\gamma \in (\hspace{0.3mm}0,1)$ is a user-chosen {\em risk} probability, and $\nu(I)$ is the {\em reliability} (coverage probability) of the confidence band. Observe that the reliability of the confidence region should be guaranteed {\em simultaneously} for all possible inputs, $x \in \CD$.

\section{Confidence Bands with Known Norm Bounds}
\label{sec:construction}
In this section, we describe the algorithm which constructs the {\em point estimates} with the corresponding {\em confidence intervals} for each query input. Then, the algorithm can be applied to estimate the intensities of missing pixels, or to construct a higher resolution image by estimating the function values on a refined grid. The starting point of our construction is the method  of \citep{csaji2022nonparametric}, which we apply to image processing problems, and later improve it, to reduce its computational complexity.

In this section, we make the simplifying assumption that a priori known bounds are available for the kernel norm of the data generating function. Later, in Section \ref{sec:kernelnorm}, we will discuss how to estimate these bounds based on the image itself, assuming that the data generating function is band-limited, i.e., it is from a Paley-Wiener space.

Our approach has similarities with the classical Wiener filter method \citep{acharya2005image} in image reconstruction, e.g., both concern with $\mathcal{L}^2$ model fitting. However, (i) our algorithm builds on {\em interpolation} instead of regression; (ii) our method is {\em nonparametric}; and most importantly, the main advantage of our approach is that (iii) it provides {\em non-asymptotically guaranteed confidence regions} for the true values.

We start with building the point estimate $\bar{f}$. The element from $\CH$ which {\em interpolates} every $y_k$ 
at the corresponding $x_k$ 
and has the {\em smallest 
norm} among such interpolants, 
\vspace{-0.5mm}
\begin{equation*}
 \bar{f} \,\doteq \, \argmin \big\{\,\|\hspace{0.3mm}f\hspace{0.4mm}\|_{\mathcal{H}}\, \mid\, f \in \mathcal{H}\hspace{1.5mm} \&\hspace{1.5mm}\forall\hspace{0.3mm} k \in [n]: f(x_k) =\, y_k   \,  \big\},
\end{equation*}
exists and for every input $x$, it takes the following form:
\vspace{-1mm}
\begin{equation}
\label{eq:minnormint}
\bar{f}(x)\, =\, \sum_{k=1}^n \hat{\alpha}_k k(x,x_k),\vspace{-0.5mm}
\end{equation}
where the weights are $\hat{\alpha} = K^{-1} y$ with $y \doteq (y_1,...,y_n)\tr$ and $\hat{\alpha} \defeq (\hat{\alpha}_1,...,\hat{\alpha}_n)$, and $K_{i,j} = k(x_i, x_j)$ is the Gram matrix \citep{paulsen2016introduction}. Since the kernel is continuous and universal by assumption A\ref{A0}, it is strictly positive definite. Moreover, we also assumed that the inputs $\{ x_k \}$ are distinct (which can be interpreted as there are no duplicated pixel intensities), therefore, matrix $K$ is always invertible in our case.

The main reason why we use \eqref{eq:minnormint} as our function estimate is that the kernel norm in an RKHS acts as a Lipsthitz-style {\em smoothness measure} \citep{paulsen2016introduction}, hence, we choose the ``smoothest'' possible function which is consistent with our observations. Note that, under assumption A\ref{A0}, $\bar{f}$ always {\em exists} and it is {\em unique}.

In this section we discuss the confidence region construction under the (auxiliary) assumption that stochastically guaranteed bounds are available for the kernel norm:
\vspace{-2mm}
\begin{assumption_b}
\label{B1} {\em For the given risk probability $\gamma \in (\hspace{0.3mm}0.1)$, we known a constant $\kappa \geq \|\hspace{0.3mm}\bar{f}\hspace{0.4mm}\|_{\CH}^2$ with}
\vspace{2mm}
$$
\mathbb{P}\big(\norm{f_*}_{\CH}^2 \leq \kappa \hspace{0.3mm}\big) \, \geq \, 1-\gamma.
$$
\end{assumption_b}
\vspace{-2mm}

A natural question which arises is that where can such bounds come from? In the context of image processing, if we are working with specific kinds of images (such as x-ray or satellite images), we could compute the kernel norms of a large number of such images and then construct $\kappa$ based on the {\em empirical quantiles} of these norms. For example, if we have the norms of $1000$ images and $
\gamma = 0.05$, then $\kappa$ can be the smallest number such that it is larger than the norms of at least $950$ images. Alternatively, we can try to build an {\em upper confidence bound} of the kernel norm based on the image itself. In Section \ref{sec:kernelnorm} we will follow this latter idea, assuming that the data generating function is band-limited as well as the inputs are independent and uniformly distributed.

The main idea of the algorithm is the following: to test a ``candidate'' $(x_0,y_0) \in \CD \times [-1, 1\hspace{0.3mm}]$ input-output pair, we compute the minimum norm which is needed to interpolate the original $\{ (x_k, y_k) \}_{k=1}^n$ dataset extended with $(x_0,y_0) \in \CD \times [-1, 1\hspace{0.3mm}]$. The {\em minimum norm interpolant} of $(x_0, y_0), \dots, (x_n, y_n)$ is now
$\tilde{f}(x)\,=\, \sum_{k=0}^n \tilde{\alpha}_k k(x, x_k)$,
where the weights are 
$\tilde{\alpha} = \gramnplus^{-1} \tilde{y}$ with $\tilde{y} \doteq (y_0, y_1, \dots, y_n)\tr\!\!,\; \tilde{\alpha} \doteq (\tilde{\alpha}_0, \dots, \tilde{\alpha}_n)\tr\!\!,$ and $K_0(i+1,j+1) = k(x_i, x_j)$ is called the {\em extended} kernel (or Gram) matrix.

Since $\mathcal{H}$ is an RKHS, we can compute the norm square of this interpolant $\tilde{f}$ as
$$
\|\hspace{0.3mm}\tilde{f}\hspace{0.4mm}\|_{\CH}^2 \,=\, \tilde{\alpha}\tr\hspace{-0.3mm} \gramnplus \tilde{\alpha}\, =\, \tilde{y}\tr\hspace{-0.3mm} \gramnplus^{-1} \gramnplus \gramnplus^{-1} \tilde{y}\,=\, \tilde{y}\tr\hspace{-0.3mm} \gramnplus^{-1} \tilde{y}.
$$

Based on these, a {\em hypothesis test} can be defined for testing the null hypothesis $H_0:y_0 = f_*(x_0)$ versus the alternative hypothesis $H_1:y_0 \neq f_*(x_0)$, as follows:
\begin{enumerate}
    \item For a given $(x_0,y_0)$ 
    pair that we want to test, first, we calculate the norm square of the {\em minimum norm interpolation} of the extended dataset $(x_0, y_0), \dots, (x_n, y_n)$.\\[-1.5mm]
    \item The null hypothesis is then accepted for the $(x_0,y_0)$ pair (in other words, it is included in the corresponding confidence band) {\em if and only if} the norm square of this (minimum norm) interpolant is less than or equal to our bound $\kappa$.
\end{enumerate}

In order to obtain the {\em endpoints} of the interval for a query input $x_0$, we have to calculate the highest and lowest $y_0$ values, which can be interpolated with a function from $\CH$ having at most norm square $\kappa$. This leads to the following two tasks:
\begin{equation}
\label{noiseless-opt-min-max}
\begin{split}
\mbox{min\,/\,max} &\quad y_{0} \\[1mm]
\mbox{subject to} &\quad (y_0, y\tr)  \gramnplus^{-1} (y_0, y\tr)\tr \leq\, \kappa,\\[0.5mm]
\end{split}
\end{equation}
where ``min\,/\,max'' means that we have to solve the problem as a minimization and also as a maximization (separately). The problems given in \eqref{noiseless-opt-min-max} are {\em convex} and they are always {\em feasible} under A\ref{A0} and B\ref{B1}, since $y^*_0 \doteq f_*(x_0)$ always satisfies the {\blue constraints.} Furthermore, they have {\em analytic} solutions discussed in \citep{csaji2022nonparametric}. 
By denoting the optimal values by $y_{\mathrm{min}}$ and $y_{\mathrm{max}}$, respectively, the endpoints of the confidence interval for $f_*(x_0)$ are given by $I_1(x_0) \defeq y_{\mathrm{min}}$ and $I_2(x_0) \defeq y_{\mathrm{max}}$. 

The pseudocode of the method, called {\em Simultaneously Guaranteed Kernel Interpolation} (SGKI), is given by Algorithm \ref{algorithm-1}. It  also describes how to construct the solutions of the (minimization and maximization) problems \eqref{noiseless-opt-min-max}, for a given query input $x_0$. 

\floatstyle{spaceruled}
\restylefloat{algorithm}

\begin{algorithm}[t!]
\caption{Simultaneously Guaranteed Kernel Interpolation (SGKI)}
\vspace{0.5mm}
\hspace*{\algorithmicindent}\textbf{Input:} dataset of observed pixels, $(x_1,y_1), \dots, (x_n,y_n)$, \\
\hspace*{\algorithmicindent}input query point $x_0 \in \CD$,
upper bound $\kappa$ for $\|f_*\|^2_\CH$. \\[1mm]
\hspace*{\algorithmicindent}\textbf{Output:} point estimate $\bar{f}(x_0)$ for the pixel value $f_*(x_0)$\\
\hspace*{\algorithmicindent}together with a confidence interval $[\hspace{0.3mm}I_1(x_0),I_2(x_0)\hspace{0.3mm}]$.
\vspace{1.5mm}
\hrule
\vspace{0.5mm}
\begin{algorithmic}[1]
\State In case $x_0 = x_k$ for any $k \in [n]$, return $\bar{f}(x_0)=y_k$ with uncertainty bounds $I_1(x_0) = I_2(x_0) = y_k$. Otherwise:
\State Calculate the minimum norm interpolant at $x_0:$\vspace{-2mm}
$$\bar{f}(x_0) = \sum_{k=1}^n \hat{\alpha}_k k(x_0,x_k),\vspace{-2mm}$$
where the weights are $\hat{\alpha} = K^{-1} y$ with $y \doteq (y_1,...,y_n)\tr$
\State Create the extended Gram matrix: for $i,j=0,\dots, n$ let
\vspace{-1.5mm}
$$K_0(i+1,j+1) \defeq k(x_i,x_j).\vspace{-1.5mm}$$
\State Compute $K_0^{-1}$, {\blue e.g., using Schur complements, see \eqref{Schur-complement-k0-inv-2},} and partition it as
\vspace{-2mm}
$$
\begin{bmatrix}
\,c & b \tr \\
\,b & A\,
\end{bmatrix}
\defeq
K_0^{-1}.\vspace{-2mm}$$
\State Let $a_0 \defeq c$, $b_0 \defeq 2\hspace{0.3mm} b \tr y$ and $c_0 \defeq y \tr A y - (n+1) \cdot \kappa$. Then, solve the quadratic equation (where we have only one variable: $y_0$) given by $a_0\, y_0^2 + b_0\, y_0 + c_0 = 0$.
\State Return $\bar{f}(x_0)$ with $I_1(x_0) \defeq y_{\mathrm{min}}$, and $I_2(x_0) \defeq y_{\mathrm{max}}$, where $y_{\mathrm{min}} \leq y_{\mathrm{max}}$ are the two solutions of the quadratic equation above (they exist, but they can coincide).
\end{algorithmic}
\label{algorithm-1}
\end{algorithm}

The SGKI method builds simultaneously guaranteed confidence bands, that is

\vspace{-1mm}
\begin{theorem}
\label{noise-free-thm}
Assume A\ref{A0} and B\ref{B1}. Then, for any risk probability $\gamma \in (0,1)$ and any finite sample size $n$, the SGKI confidence band is guaranteed to have reliability\,
$\nu(I) \, \geq \, 1-\gamma\vspace*{0.8mm}.\vspace*{-1mm}$
\end{theorem}
\vspace{-1mm}
The proof basically follows from the construction itself. That is, under our assumptions, we have $\mathbb{P}( \norm{f_*}_\mathcal{H}^2 \leq \kappa ) \,\geq\, 1-\gamma$. In case $\norm{f_*}_\mathcal{H}^2 \leq \kappa$, then for all $x_0$, $f_*(x_0)$ is in the confidence band, since $f_*$ interpolates $\{(x_k, f_*(x_k))\}_{k=0}^n$, and its norm is $\leq \kappa$, thus the minimum norm interpolants of these extended datasets inherit this norm bound. The reader is referred to \citep{csaji2022nonparametric,csaji2023eysm} for more details of the method and the fundamental ideas behind the construction.

Besides the {\em non-asymptotic} coverage guarantee above, we should also show that the estimate (minimum norm interpolant) is always included in the confidence band:
\vspace{-3mm}
\begin{lemma}
{\em If A\ref{A0}, B\ref{B1}, then $\forall\hspace{0.3mm} x\in\CD: \bar{f}(x) \in  [\hspace{0.3mm}(I_1(x), I_2(x)\hspace{0.3mm}]$.}
\end{lemma}
\vspace{-5mm}
\begin{proof}
Let us fix a query input $x_0 \in \CD$. If there is a $k \in [n]$, such that $x_0 = x_k$, then the statement trivially holds.
Now assume that $x_0 \neq x_k$ for all $k \in [n]$. Let us introduce:
\begin{equation*}
\CG_{y}^{\hspace{0.2mm}\kappa}\; \defeq\, \big\{\, f \in \CH \,\mid\, \forall\hspace{0.3mm} k \in [n]: f(x_k) = y_k,
f(x_0) = y, \; \|\hspace{0.3mm}f\hspace{0.4mm}\|_{\mathcal{H}}^2 \leq \kappa \, \big\},    
\vspace{1mm}
\end{equation*}
which contains those interploants of the extended dataset that has a norm square less than or equal to our upper bound $\kappa$.
We can observe that $\bar{f} \in \CG_{\bar{f}(x_0)}^{\hspace{0.2mm}\kappa}$, since $\kappa \geq \|\hspace{0.3mm}\bar{f}\hspace{0.4mm}\|_{\CH}^2$. Then,
$$
I_1(x_0)\, = \min_{y\,:\,\CG_{y}^{\hspace{0.2mm}\kappa}\, \neq\, \emptyset}\, y\, \leq\, \bar{f}(x_0)\, \leq\, \max_{y\,:\,\CG_{y}^{\hspace{0.2mm}\kappa}\, \neq\, \emptyset}\, y\, =\, I_2(x_0),
$$
which proves the statement of the lemma, as the query input $x_0\in\CD$ was arbitrary.
\end{proof}

Note that the confidence interval construction of Algorithm \ref{algorithm-1} could also be used together with other image restoration methods, such as Total Variation (TV). It provides a general approach for uncertainty quantification, even without $\bar{f}$.

The standard form of SGKI works with grayscale images, as we assume that $f_*$ is scalar-valued. 
In case of {\em color} images, the pixels are given by multiple numbers, e.g., by RGB or CMYK codes. In this case, $f_*: \CD \to [-1, 1\hspace{0.3mm}]^m$ for some $m \in \mathbb{N}$, typically $m=3$ or $m=4$. Thus we have {\em vector-valued} outputs. The simplest way to handle this issue is to estimate the true value of each coordinate separately. Then, for each query input $x_0 \in \CD$ 
we have a confidence interval for each possible coordinate. These intervals can be combined into a {\em hyperrectangle} to get a confidence region for $f_*(x_0)$.

Up to this point, we can apply any RKHSs with a universal kernel. Paley-Wiener spaces become crucial for getting stochastic upper bounds for the kernel norm.

\section{Bounding the Kernel Norm}
\label{sec:kernelnorm}
Now, we address the problem of computing $\kappa$, i.e.\ a stochastic upper bound of $\|f_*\|_{\CH}^2$ from the image, to get rid of B\ref{B1}. In order to do so, we introduce three new assumptions. One of our new assumptions is that the data generating function underlying our observations is band-limited, which is a standard setup for signal processing.

\vspace{-2mm}
\begin{assumption}
\label{A1} 
{\em Function $f_*$ is from a Paley-Wiener space $\CH$; and $f_*$ is almost time-limited to 
$\CD$, that is
\vspace{1mm}
\begin{equation*}
\int_{\mathbb{R}} f^2_*(x)\,\BI(x \notin  \CD) \: \lambda(\mathrm{d}x) \, \leq \, \delta_0,
\vspace{1mm}
\end{equation*}
where $\BI(\cdot)$ is an indicator and $\delta_0 > 0$ is a universal constant.}
\end{assumption}
\vspace{-2mm}

Assumption A\ref{A1} limits the frequency domain of $f_*$, hence, its Fourier transform cannot have arbitrarily high frequencies, that is, the change of $f_*$ cannot be arbitrarily fast. This is needed to restrict the model class and to ensure that we can generalize well to out-of-sample inputs. Because of the {\em Fourier uncertainty principle},
we must allow the ``true'' function to be defined outside of $\CD$, but this part of $f_*$ 
should be ``negligible'', i.e., its norm cannot exceed a (known) small constant, $\delta_0$. 

\vspace{-2mm}
\begin{assumption}
\label{A2} 
{\em Sample $(x_1, y_1), \dots, (x_n, y_n) \in \CD \times \RR$ is independent and identically distributed $($i.i.d.$)$.}
\end{assumption}
\vspace{-10mm}
\begin{assumption}
\label{A3} {\em The inputs, $\{x_k\}$, 
are distributed uniformly on $\CD$.}
\end{assumption}
\vspace{-2mm}

Then, we can construct $\kappa$ based on the ideas of \citep{csaji2022nonparametric}:
\vspace{-2mm}

\begin{lemma}
\label{lemma-noise-free-norm-estimation}
{\em Assuming A\ref{A1}, A\ref{A2}, and A\ref{A3}, for any risk probability $\gamma \in (0,1)$, we have
\vspace{1mm}
$$
\mathbb{P}\big(\norm{f_*}_{\CH}^2 \leq \kappa \hspace{0.3mm}\big) \, \geq \, 1-\gamma,
\vspace{1mm}
$$
with the following upper bound:
\vspace{1mm}
\begin{equation}
\label{eq:kappa}
\kappa \, \defeq\, \frac{1}{n} \sum_{k=1}^n y_k^2 + \sqrt{\frac{\ln(\gamma)}{-2n}} + 
\delta_0.
\vspace{-1mm}
\end{equation}}
\end{lemma}

This result was extended in \citep{csaji2023eysm}, using importance sampling, to the case of having arbitrary a priori known continuous input distributions, but for simplicity, here we kept the uniformity assumption of the inputs in this paper.    

In image reconstruction, we often work with {\em quantized inputs}, i.e., $f_*$ cannot be queried at arbitrary points, but there is a {\em lattice structure} $\{A_i\}_{i=1}^r$ induced by the points of a {\em sampling grid} $\{\bar{x}_i\}_{i=1}^r$, where $r$ is the spatial resolution \citep{acharya2005image}. We can extend the result of Lemma \ref{lemma-noise-free-norm-estimation} by taking into account the quantization error of inputs in \eqref{eq:kappa}, namely, by using $\kappa_r \defeq \kappa + \delta_r$, where
$$
\delta_r \defeq \max_{i \in [r]} \sup_{x\in A_i} |f^2_*(\bar{x}_i) - f^2_*(x)|,
$$
assuming this local Lipschitz constant, or an upper bound of it, is known.
Fortunately, $\delta_r$ typically becomes negligible as the resolution increases and the sampling approaches the {\em Nyquist rate} \citep{acharya2005image}. Therefore, we can typically assume that we sample the pixels themselves using a discrete uniform distribution.

\section{Reducing the Computational Complexity}
\label{sec:schur}
In this section, we improve the {\em computational efficiency} of the SGKI algorithm by studying the solutions of optimization problems \eqref{noiseless-opt-min-max}. Even though these convex problems can be solved analytically, their solutions include the calculation of $K_0^{-1}$ for every query input $x_0 \in \CD$,
as $K_0$ also depends on $x_0$. Computing the inverse 
of $K_0$ for each missing pixel could become computationally demanding.
Here, we propose an alternative, more efficient solution by using the concept of Schur complements.

The core idea is that $K_0 \in \mathbb{R}^{(n+1) \times (n+1)}$ is essentially the original $K \in \mathbb{R}^{n \times n}$ extended by the values induced by query input $x_0$. We argue that we only need to compute matrix $K^{-1}$ once, and then we can use it to compute $K_0^{-1}$ for every $x_0$.

Formally, for a given query input $x_0$, $K_0$ can be written as a block matrix,
$$
K_0 = 
\begin{bmatrix}
r_0 & k_0\, \\
\,k_0 \tr & K
\end{bmatrix}\!,
$$
where $K \in \mathbb{R}^{n \times n}$, $k_0 \in \mathbb{R}^n, r_0 \in \mathbb{R}$, with $r_0 = k(x_0, x_0)$ and $k_{0,i} = k(x_0, x_i),$ for $i \in [n]$.

Now, let us introduce the {\em Schur complement}
$$g_0 \defeq (K_0/r_0) \defeq r_0 - k_0 \tr K^{-1} k_0,$$
then, we can calculate $K_0^{-1}$ by exploiting the fact that \citep{boyd2004convex},
\begin{equation}
\label{Schur-complement-k0-inv-2}
K_0^{-1} =
\begin{bmatrix}
g_0^{-1} & -K^{-1} k_0 g_0^{-1} \\
\,-g_0^{-1} k_0 \tr K^{-1}\quad & K^{-1} + K^{-1} k_0 g_0^{-1} k_0 \tr K^{-1}\,
\end{bmatrix}\!.
\vspace{1mm}
\end{equation}
This approach has the advantage that \eqref{Schur-complement-k0-inv-2} can be computed using $\CO(n^2)$ floating point operations (flops), instead of $\CO(n^3)$ flops, assuming $K^{-1}$ is available. Consequently, we only need to compute the inverse of matrix $K$ once, as an initialization step, and then matrix $K^{-1}$ can be used for each possible query input $x_0$ to construct the inverse of the corresponding extended kernel matrix $K_0$. For a given $x_0$, first we need to compute the query point dependent scalar $r_0$ and vector $k_0$, based on which we can calculate the Schur complement $g_0$. Finally, we can apply formula \eqref{Schur-complement-k0-inv-2} to build $K_0^{-1}$ in a computationally efficient way. {\blue For example, if $n=100$, then the speedup of computing the inverse of $K_0$, given $K^{-1}$, could be even $100 \times$, depending on the implementation.}

{\blue Taking the whole image into account, the computational complexity of the methods are as follows. Assume that the resolution of the image is $h \times w$ pixels and we observe $n$ of them. Then, the (parameterized) complexity of the original method is $\CO((hw-n)n^3)$ flops, while the Schur complement based version requires $\CO(n^3+(hw-n)n^2))$ flops. The first term corresponds to computing the inverse of matrix $K$ in the initial phase.}

\section{Empirical Validation}
\label{sec:experiments}

In this section we present a series of {numerical experiments} with the proposed method, to demonstrate its performance. We have investigated both {\em synthetic} (artificially generated) and {\em real-world images} and studied both {\em inpainting} and {\em super-resolution} problems. One of the main drawcards of simultaneously guaranteed kernel interpolation (SGKI) is its non-asymptotic uncertainty quantification (UQ), but since most image processing methods do not offer UQ, it is hard to fairly compare SGKI with other methods, apart from only using $\bar{f}$, the associated SGKI point estimate.

For {\em quantitative} comparisons, we used three quality metrics which we now introduce for the sake of completeness. They measure specific ``distances'' between the original image and the reconstructed one. Let us start by recalling the standard {\em mean squared error} (MSE) criterion for matrices $A, B \in \mathbb{R}^{h \times w}$, where $h$ and $w$ denote the height and the width of the matrix (e.g., encoding a grayscale image), respectively: 
\vspace{-1mm}
\begin{equation*}
    \mbox{MSE}(A,B) \,\defeq\, \frac{1}{h \, w}\, \| A - B\hspace{0.3mm}\|^2_{\scriptscriptstyle{\text{F}}} \,=\, \frac{1}{h \, w}\, \sum_{i=1}^{h} \sum_{j=1}^w\, (A_{i,j}-B_{i,j})^2\!\!,
    \vspace{-1mm}
\end{equation*}
where $\|\cdot\|_{\scriptscriptstyle{\text{F}}}$ denotes the {\em Frobenius norm}. The three {\em quality metrics} were as follows.
\begin{itemize}
    \item {\em Peak Signal-to-Noise Ratio} (PSNR) is often defined in signal processing as the logarithm of the ratio between the ``power'' of the noise (typically using the decibel scale) and the maximum possible ``power'' of the signal. In our case, it is defined as
    \vspace{-1mm}
    \begin{equation*}
        \mbox{PSNR}(A,B)\, \defeq\, 10\, \log_{10} \Big( \frac{M^2}{\mbox{MSE}(A,B)}\Big),
    \vspace{0mm}
    \end{equation*}
    where $M$ is the maximum pixel value (e.g., for grayscale images it is usually $255$).\\[-2mm]
    \item {\em Structural Similarity Index Measure} (SSIM) quantifies the similarity between two images by taking several parameters into account, such as contrast, luminance and structure \citep{wang2003multiscale}. Its formula is given by
    \begin{equation*}
        \mbox{SSIM}(A,B)\, \defeq\, \frac{(2 \mu_A \mu_B + c_1)(2 \sigma_{AB}+c_2)}{(\mu_A^2+\mu_B^2+c_1)(\sigma_A^2+\sigma_B^2+c_2)},
    \end{equation*}
    where $\mu_A$ and $\mu_B$ denote the sample mean, $\sigma_A^2$ and $\sigma_B^2$ are the (empirical) variances, $\sigma_{AB}$ is the (empirical) covariance of the pixels for images $A$ and $B$ and $c_1 \defeq (k_1 L)^2$, $c_2 \defeq (k_2 L)^2$ are variables, which stabilize the division. $L$ is the dynamic range of the pixel values, while $k_1 = 0.01$ and $k_2 = 0.03$ are constant parameters.\\[-2mm]
    
    \item {\em Normalized Root Mean Square Error} (NRMSE) is a normalized version of the root of the mean squared error (RMSE). In our case, we used the following normalization
    \begin{equation*}
        \mbox{NRMSE}(A,B)\, \defeq\, \frac{\sqrt{\mbox{MSE}(A,B)} \cdot \sqrt{h \, w}} {\norm{A\hspace{0.3mm}}_{\scriptscriptstyle{\text{F}}}}\,=\, \frac{\norm{A-B\hspace{0.3mm}}_{\scriptscriptstyle{\text{F}}}}{\norm{A\hspace{0.3mm}}_{\scriptscriptstyle{\text{F}}}},
    \end{equation*}
    where $A$ plays the role of the original ``true'' image and $B$ is the reconstructed one.
\end{itemize}

\subsection{Synthetic Test Images}
\label{subsec:synt-test}
We start by presenting our experiments on {\em synthetic} test images. We generated $100$ images based on data generating functions that were guaranteed to be contained in our chosen $\CH$. We used a Paley-Wiener RKHS with parameter $\eta = 50$ and the ``true'' function was constructed as follows: $20$ random $\bar{x}_k \in [0,1]^2$, $k \in [20]$ knot points were generated with uniform $[\hspace{0.3mm}0,1]$ distribution on both of their coordinates. Then $f_*(x) = \sum_{k=1}^{20} w_k k(x, \bar{x}_k)$ was created, where each $w_k$ had a uniform distribution on $[-1,1]$. The obtained function was normalized, in case its maximum exceeded $1$. The resolution of the images was $r \times r$ with $r=50$. For the pixel at position $(i,j)$, where $i, j \in [50]$, we have $y_{i,j} \defeq f_*(x_k)$, if pixel $(i, j)$ is observed and its input vector (i.e., on the sampling grid) is $x_k \in \CD \defeq [\hspace{0.5mm}0, 1\hspace{0.3mm}]^2\!\!$, e.g., $x_{k,1} = i/(r+1)$  and $x_{k,2} = j/(r+1)$.

{\renewcommand{\arraystretch}{1.6}
\begin{table}[!b]
\caption{Synthetic single image inpainting experiments using $100$ randomly generated ``band-limited'' images with $10\,\%$ observations.}
\begin{tabular}{|l||c|c|c|}
\hline
Inpainting        & PSNR (avg)              & SSIM (avg)            & NRMSE (avg)           \\ \hline\hline
SGKI-PW   & \textbf{26.0489} & \textbf{0.7499} & 0.1082 \\ \hline
Total Variation   & 23.3874  & 0.5354 & 0.1356 \\ \hline
Biharmonic   & 25.4460 & 0.6872 & \textbf{0.1074} \\ \hline
Large Mask   & 22.6320 & 0.4711 & 0.1476 \\ \hline
\end{tabular}
\label{tab1}
\end{table}
}

For the single image inpainting problem, just (random) $10\,\%$ of the pixels of the generated images were observed. The interpolant of SGKI, with the PW kernel, was compared with three different methods: {\em Total Variation} \citep{combettes2004image}, which estimates the values of the missing pixels by minimizing the total variation subject to the matching known pixel values; an algorithm utilizing the {\em biharmonic equation} \citep{damelin2018surface}; and {\em Large Mask Inpainting} \citep{suvorov2022resolution}, which uses fast Fourier convolutions. 
The results of the quantitative experiments are summarized in Table \ref{tab1}, where the averages of PSNR, SSIM and NRMSE are shown.

Note that for the PSNR and the SSIM quality metrics, higher values indicate better reconstruction performance, while for NRMSE, lower values are preferred. 
{\blue The best result for each case in the comparison tables is highlighted using bold fonts.}

This experiment shows that if the characteristics of the image fit perfectly our {\em inductive hypothesis}, induced by our kernel choice (which was ensured by our image generation process), then SGKI provides an excellent performance and outperforms standard, widespread inpainting methods (w.r.t.\ the PSNR and SSIM metrics; for NRMSE, the biharmonic approach was slightly, but not significantly better).

\begin{figure}[!t]
    \centering
	\hspace*{-2mm}		
	\includegraphics[width = \columnwidth]{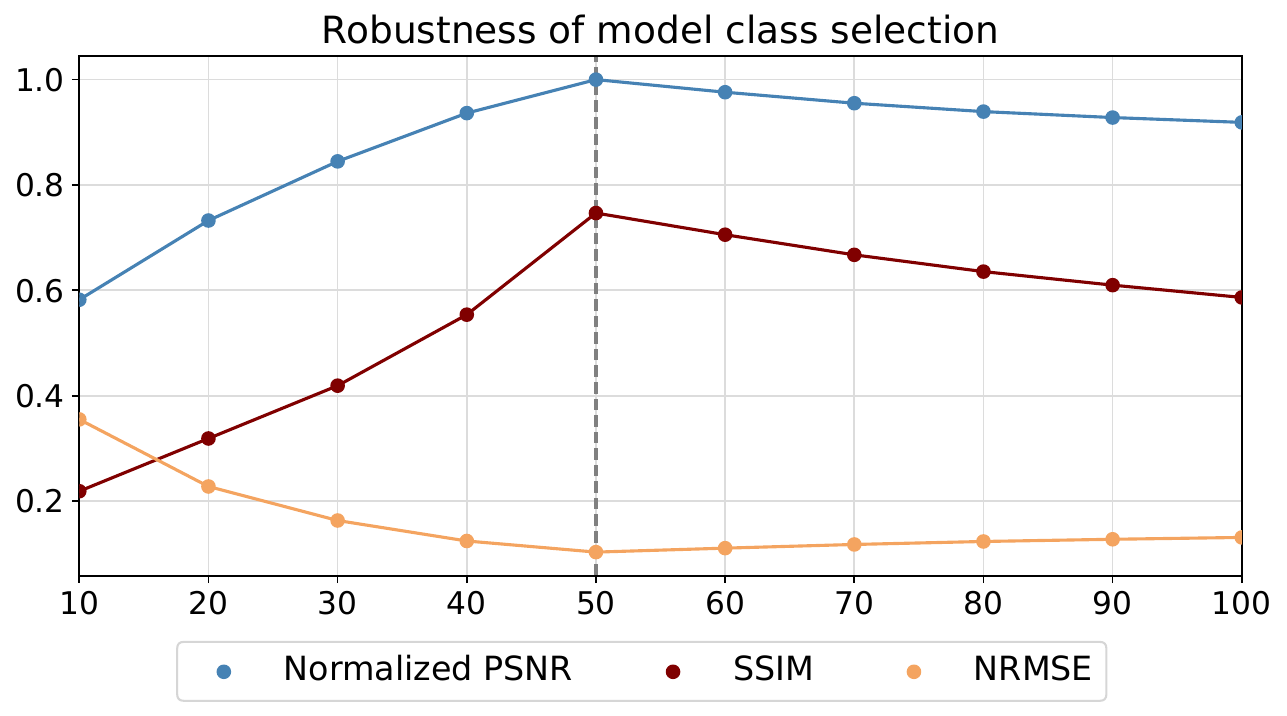} 	
    \caption{Synthetic single image inpainting experiment, with $10 \%$ of the pixels observed, about the robustness of model class selection. The $x$-axis represents the different $\eta$ (hyper-parameter) values of the applied Paley-Wiener kernel, while the $y$-axis shows the respective (averaged) values of the three image quality metrics (PSNR was normalized by its highest realization, in order to get its values in $[0,1]$). The synthetic images were generated with the ``true'' value $\eta_*=50$, which is marked by a gray dashed line.}
\label{fig:experiment1}
\end{figure}

We also investigated how {\em robust} the SGKI method is w.r.t.\ misspecified model classes. In order to test this, 
we used the same $100$ synthetic images as before with random $10\,\%$ of the pixels as inputs. We had $\eta_* = 50$, while we used other $\eta$ (hyper-parameter) values for our PW kernel. Recall that $\eta_*$ specifies the bound on the allowed frequencies of the underlying ``true'' data-generating function. The resulting (averaged) quality metrics ($y$-axis) depending on the chosen kernel hyper-parameter ($x$-axis) are shown in Figure \ref{fig:experiment1}. The results demonstrate that choosing too low $\eta$ values (i.e., lower than the true frequency bound) could result in significant performance losses, but SGKI is much more robust against over-bounding the frequencies, as increasing $\eta$ above the ``true'' $\eta_*=50$ typically only results in very modest performance losses.

\begin{figure}[!ht]
    \centering
	\hspace*{-2mm}		
	\includegraphics[width = 0.91\columnwidth]{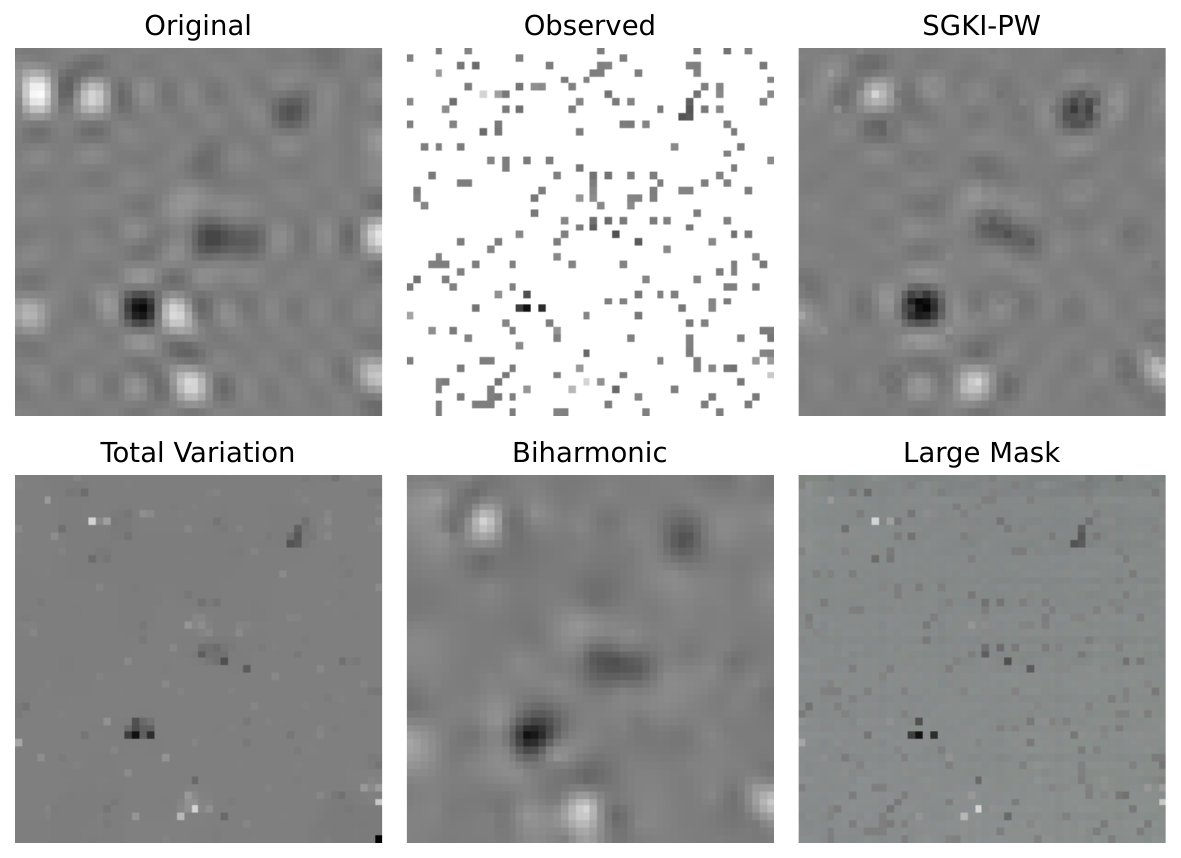} 	
    \caption{Synthetic single image inpainting experiment with random $10 \%$ of the pixels observed.}
\label{fig:experiment2}
\vspace*{-1mm}
\end{figure}

\begin{figure}[!ht]
    \centering
	\hspace*{-2mm}		
	\includegraphics[width = 0.91\columnwidth]{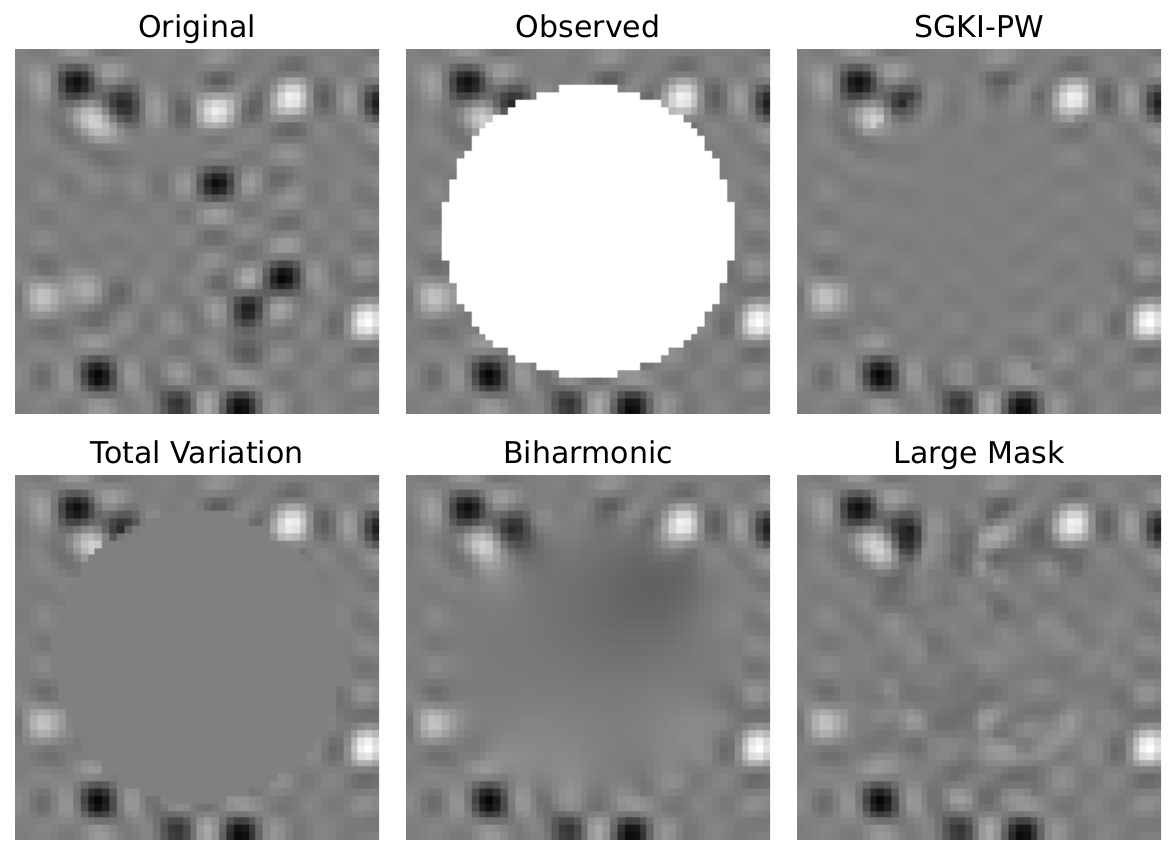} 	
    \caption{Synthetic single image inpainting experiment with a large, deterministic cut-out: instead of random pixels, a circle was removed. In this case, $1244$ from the $2500$ the pixels were observed.}
\label{fig:experiment3}
\vspace*{-2.5mm}
\end{figure}

Figures \ref{fig:experiment2} and \ref{fig:experiment3} illustrate the visual quality of the obtained results. Specifically, for Figure \ref{fig:experiment3} non-random observations were used, as a filled circle shape was cut out from the image. In this latter case, the Large Mask approach (PSNR: $24.9824$, SSIM: $0.6298$, NRMSE: $0.1122$) performed similarly to SGKI (PSNR: $25.0751$, SSIM: $0.6747$, NRMSE: $0.1110$). On the other hand, Large Mask provided a poor result for the case of random observations, for which the Biharmonic method had similar performance to SGKI. Conversely, the Biharmonic method did a poorer job for a large cut-out. In conclusion, for synthetic images (for which we can guarantee that our assumptions are satisfied), SGKI provided the most stable results for the inpainting problem.

Regarding the {\em super-resolution} problem, $20$ synthetic images were generated using the same method as before. A Paley-Wiener RKHS was used with parameter $\eta = 50$. The resolution of the images was $2\hspace{0.3mm}r \times 2\hspace{0.3mm}r$ with $r = 50$, which was reduced to $r \times r$ pixels by using {\em subsampling}, namely, every second pixel in every second row (hence, $1$ out of $4$ pixels) was chosen  to create an image with resolution $r \times r$. These reduced images were used as inputs for the tested super-resolution methods, and the reconstructed images were then compared with the original ones based on our quality metrics.

SGKI was also compared with other methods in this case: two state-of-the-art approaches based on deep
learning, namely {\em Enhanced Deep Super-Resolution} 
\citep[EDSR,][]{Lim_2017_CVPR_Workshops}, and {\em Multi-Scale Residual Network} \citep[MSRN,][]{li2018multi}.
Next to that, we have used several other methods in the comparison, as well, which provide their estimates based on nearby pixel value(s): {\em Bicubic Interpolation} \citep{keys1981cubic}, which uses the $4 \times 4$ neighborhood, and {\em Bilinear Interpolation} \citep{fadnavis2014image}, which uses a $2 \times 2$ neighborhood while determining the value of a new pixel. The {\em Nearest-Neighbor Interpolation} \citep{lehmann1999survey} uses the value of the nearest pixel. Meanwhile, the {\em Lanczos Resampling} \citep{madhukar2013lanczos}, which is based on the {\em sinc} function, uses a $8 \times 8$ 
\FloatBarrier 
neighborhood (in our case) for estimation. All tested super-resolution methods were used with a $\times 2$ scale. The obtained quantitative results are summarized in Table \ref{tab2}.

{\renewcommand{\arraystretch}{1.6}
\begin{table}[!b]
\caption{
Synthetic single image super-resolution experiments using $20$ 
``band-limited'' images with the aim to double their resolution.}
\begin{tabular}{|l||c|c|c|}
\hline
Super-resolution         & PSNR (avg)              & SSIM (avg)            & NRMSE (avg)           \\ \hline
\hline
SGKI-PW    & \textbf{38.1143}  & \textbf{0.9831} & \textbf{0.0258} \\ \hline
EDSR     & 36.1177 & 0.9741 & 0.0311 \\ \hline
MSRN     & 36.1017 & 0.9740 & 0.0312 \\ \hline
Bicubic  & 35.9471 & 0.9734 & 0.0318 \\ \hline
Bilinear & 35.8641 & 0.9725 & 0.0321 \\ \hline
Lanczos  & 36.0334 & 0.9745 & 0.0315 \\ \hline
Nearest Neighbor & 33.2433 & 0.9447 & 0.0433 \\ \hline
\end{tabular}
\label{tab2}
\end{table}
}

It can be observed that for experiments with synthetic data, in which the images were generated in a way that they were guaranteed to satisfy our assumptions, the SGKI method provided the best results even for the super-resolution problem.

Hence, for ``band-limited'' images, SGKI provided excellent performance for both inpainting and super-resolution problems, and it was also shown to be robust.

\begin{figure}[!t]
{\bluetext
    \centering
	\hspace*{-2mm}		
	\includegraphics[width = \columnwidth]{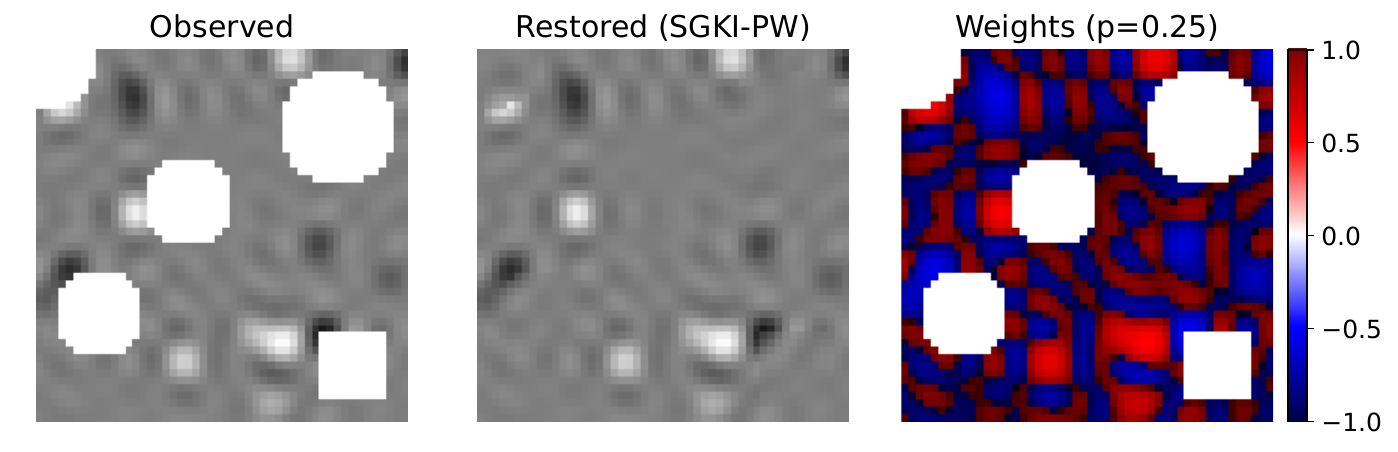} 	
    \caption{Kernel weights of the SGKI point estimate in an inpainting problem for a synthetic image.}
\label{fig:experiment-w1}
\vspace*{-1mm}
}
\end{figure}

{\blue We have studied the kernel weights of the minimum norm interpolant, as well. Recall that this interpolant,
shown in Algorithm \ref{algorithm-1}, is our point estimate of the missing pixels. 
We have generated a synthetic image, removed some parts of it, 
and plotted the kernel weights, $\{ \hat{\alpha}_k\}$, to demonstrate the impacts of the observed pixels. The weights were first rescaled to the interval $[-1,1\hspace{0.3mm}]$, then for each weight $\hat{\alpha}_k$, we plotted $\hat{\alpha}_k^p$ with $p=0.25$. We did this last step to enhance the visibility of weights closer to zero.
Figure \ref{fig:experiment-w1} illustrates the weights for a synthetic inpainting problem. This experiment demonstrates which pixels or patters are more ``informative'' with respect to the image we are processing. Note that only the available (actually observed) pixels have weights.}

{\bluetext Finally, we evaluated the computational speedup (on a standard PC with an Intel i7 processor) of using Schur complements in \eqref{Schur-complement-k0-inv-2} for Gramian inversion, assuming $K^{-1}$ is given, compared to direct inversion. Testing on random $64 \times 64$ synthetic images with varying observed pixel percentages, we found an average speedup of $12\times-13\times$, see Table \ref{tab-time}, over $1000$ trials. Each trial (a column of the table) consisted of $100$ experiments on randomly chosen pixel locations across $10$ different instances of missing pixels. This speedup is expected to increase linearly with the number of observed pixels.

{\renewcommand{\arraystretch}{1.6}
\begin{table}[!t]
{\bluetext
\caption{
Average computation times (in seconds) and speedup ratios for calculating the inverse Gram matrix using the Schur complement based method versus the standard matrix inversion,
as a function of the percentage of removed pixels in synthetic inpainting tasks.}
\begin{tabular}{|l||c|c|c|c|c|}
\hline
Pixel removal percentage  & $5\,\%$              & $10\,\%$            & $15\,\%$ & $20\,\%$ & $25\,\%$           \\ \hline
\hline
Time (standard inverse)   & $1.0999$  & $0.9963$ & $0.8314$ & $0.6560$ & $0.5745$\\ \hline
Time (Schur-based inverse) & $0.0822$ & $0.0723$ & $0.0622$ & $0.0543$ & $0.0468$\\ \hline
Speedup (standard / Schur)    & $13.3761\times$ & $13.7715\times$ & $13.3781\times$ & $12.0968\times$ & $12.3043\times$\\ \hline
\end{tabular}
\label{tab-time}
}
\end{table}
}

}

\subsection{Real-World Test Images}

We now present our experiments on {\em real-world} test images. Quantitative experiments were also performed for both problems on the (rescaled) 
{\em Set12} dataset \citep{zhang2017beyond}, which consists $12$ grayscale images with resolution $256 \times 256$. The photos show people, animals, vegetables, houses and vehicles.

For the single image inpainting task, the images were rescaled by using Bicubic interpolation. In this case, SGKI was used with the Gaussian kernel with parameter $\sigma = 0.05$; recall that this kernel is defined as $k(z,s) \defeq \exp (-||z-s||^2 / (2 \sigma^2))$. 

For the inpainting problem, $10$ separate experiments were performed for each image in the dataset, with (random) $10\,\%$ of the pixels observed. For the super-resolution problem, we again applied subsampling, but now every fourth pixel in every fourth row (thus, $1$ out of $16$ pixels) was chosen to create the rescaled images. For this task, the Paley-Wiener kernel was applied with parameter $\eta = 175$ and the super-resolution algorithms were used with scale $\times 4$. The results are summarized in Tables \ref{tab3} and \ref{tab4}.

The SGKI method, with the Gaussian kernel, provided solid results for the inpainting problem on this dataset, though the Total Variation and the Biharmonic methods were slighly better. On the other hand, SGKI with a PW kernel showed excellent results for the super-resolution problem, it achieved even better results than EDSR and MSRN. We emphasize that we applied subsampling to obtain the lower resolution images, as this technique ensured the setup we assumed regarding the observations.

{\renewcommand{\arraystretch}{1.6}
\begin{table}[!b]
\caption{Single image inpainting experiments, with $10\,\%$ pixels observed, on real-world images from the grayscale Set12 dataset.}
\begin{tabular}{|l||c|c|c|}
\hline
Inpainting        & PSNR (avg)              & SSIM (avg)            & NRMSE (avg)           \\ \hline \hline
SGKI-G   & 16.7601 & 0.3792 & 0.2765 \\ \hline
Total Variation   & 17.0084 & 0.4201 & 0.2693 \\ \hline
Biharmonic   & \textbf{17.3096} & \textbf{0.5260} & \textbf{0.2605} \\ \hline
Large Mask   & 14.7450 & 0.2237 & 0.3477 \\ \hline
\end{tabular}
\label{tab3}
\end{table}
}

\begin{figure}[!t]
    \centering
	\hspace*{-2mm}		
	\includegraphics[width = \columnwidth]{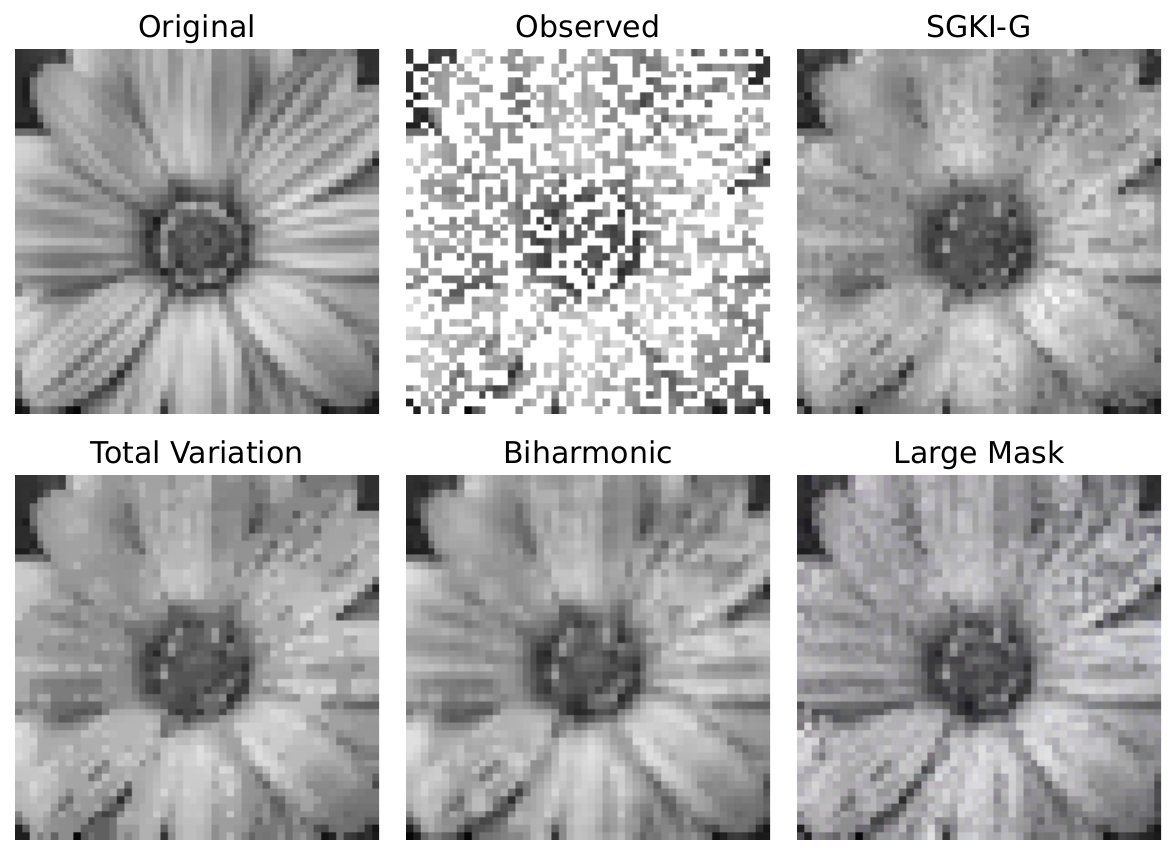} 	
    \caption{Visual comparison of single image inpainting methods with $50 \%$ of the pixels observed.}
\label{fig:experiment4}
\vspace*{-2mm}
\end{figure}
\begin{figure}[!t]
    \centering
	\hspace*{-2mm}		
	\includegraphics[width = \columnwidth]{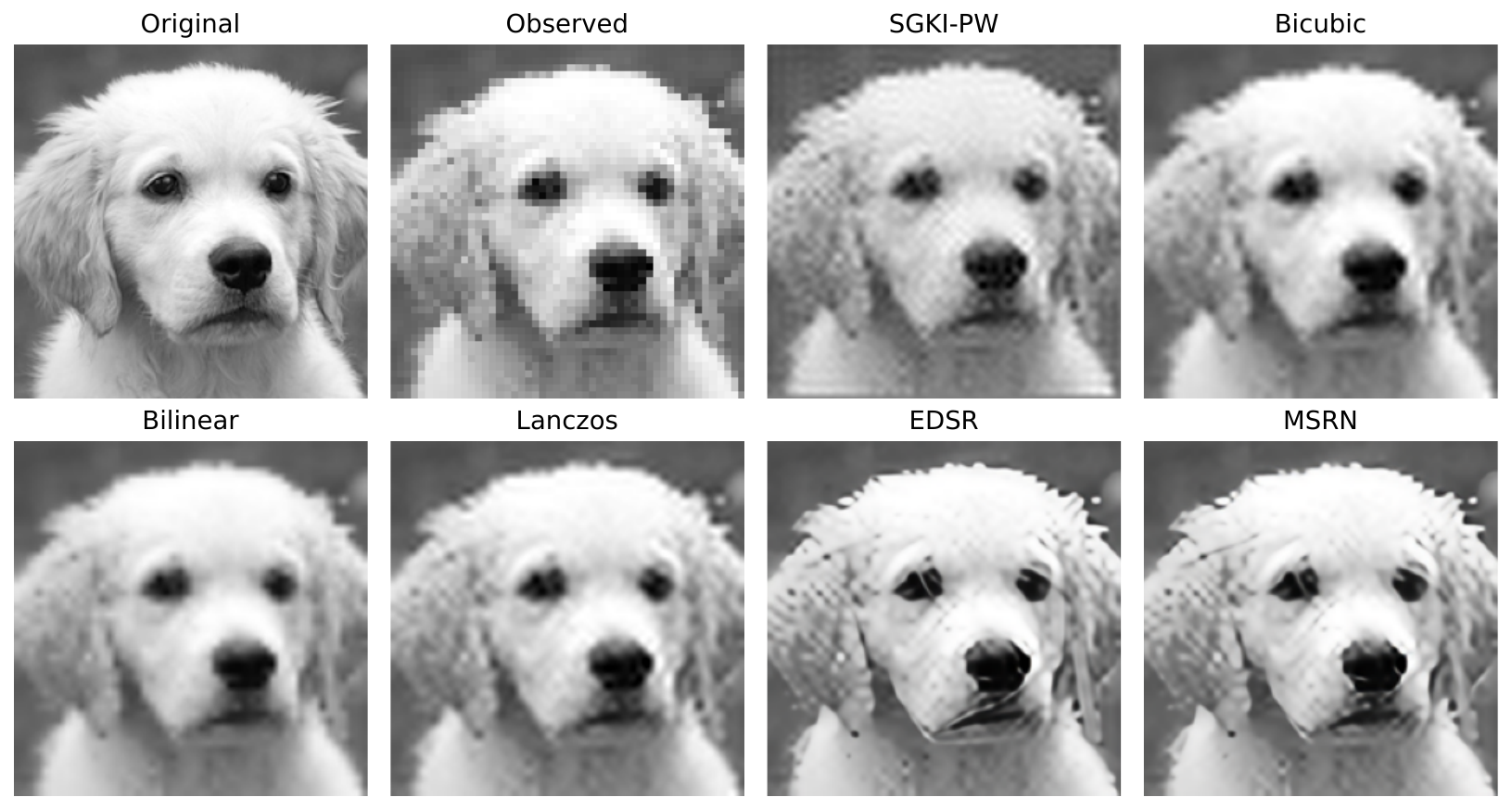} 	
    \caption{Visual comparison of single image super-resolution methods; the target scale was $\times 4$.}
\label{fig:experiment5}
\vspace*{-1mm}
\end{figure}
\FloatBarrier

{\renewcommand{\arraystretch}{1.6}
\begin{table}[!b]
\caption{Single image super-resolution experiments on real-world images from the grayscale Set12 dataset; the target scale was $\times 4$.}
\begin{tabular}{|l||c|c|c|}
\hline
Super-resolution         & PSNR (avg)              & SSIM (avg)            & NRMSE (avg)           \\ \hline\hline
SGKI-PW    & \textbf{21.6117}  & \textbf{0.6263} & \textbf{0.1597} \\ \hline
EDSR     & 16.7301  & 0.5049 & 0.2773 \\ \hline
MSRN     & 16.8671 & 0.5078 & 0.2729 \\ \hline
Bicubic  & 19.9679 & 0.6119 & 0.1911 \\ \hline
Bilinear & 20.4913 & 0.6211 & 0.1801 \\ \hline
Lanczos  & 19.8083 & 0.5967 & 0.1946 \\ \hline
Nearest Neighbor & 18.8881 & 0.5624 & 0.2166 \\ \hline
\end{tabular}
\label{tab4}
\end{table}
}

As a visual illustration of inpainting, Figure \ref{fig:experiment4} shows an experiment, where the original picture\footnote{All of the real-world photos presented in the paper were obtained from Pixabay, they were also edited.} had resolution $50 \times 50$ and only $1250$ pixels were observed from the total 2500. Here, the Gaussian kernel was used with (hyper-) parameter $\sigma = 0.03$. Label ``SGKI-G'' highlights this choice, while ``SGKI-PW'' indicates the PW kernel.

Figure \ref{fig:experiment5} presents a visual illustration for super-resolution methods. In this experiment, an image with $4\hspace{0.3mm}r \times 4\hspace{0.3mm}r = 200 \times 200$ pixels was used, which was reduced to $r \times r$ pixels $(r = 50)$ with subsampling, as before, selecting $1$ out of $16$ pixels to be kept.
In the SGKI case, the Paley-Wiener kernel was applied with parameter $\eta = 175$.
The output for the Nearest Neighbor (NN) method was not plotted in Figure \ref{fig:experiment5}, since the resulted image does not differ from the observed image except from its resolution.

It is important to note that the results regarding SGKI only show the minimum norm interpolant, $\bar{f}$,
which is basically the center of the confidence band. However, one does not need to use this interpolant, the UQ of SGKI can be combined with other methods to evaluate their uncertainty. The simultaneous SGKI confidence bands are guaranteed irrespectively of the actual method used to estimate the missing pixels.

\begin{figure}[!t]
{\bluetext
    \centering
	\hspace*{-2mm}		
	\includegraphics[width = \columnwidth]{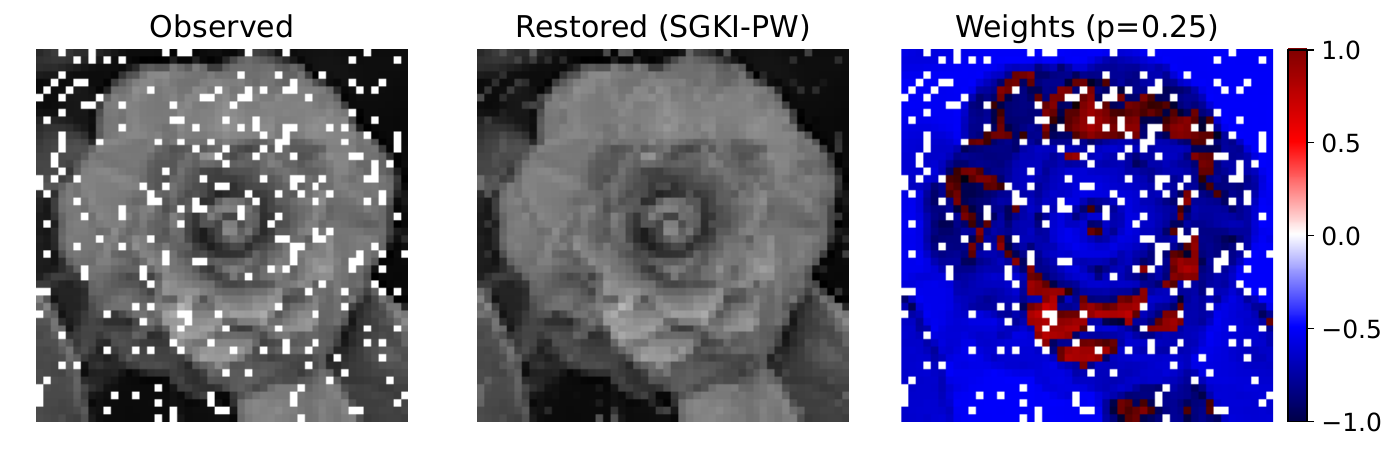} 	
    \caption{Kernel weights of the SGKI point estimate in an inpainting problem for a real-world image.}
\label{fig:experiment-w2}
}
\end{figure}

{\blue We have plotted the kernel weights of the minimum norm interpolant for a real-world image, too. In this case, random $90\,\%$ percent of the pixels were observed. The Paley-Wiener kernel was used with $\eta = 50$. 
The results are shown in Figure \ref{fig:experiment-w2}, where the weights were transformed similarly to the kernel weight experiment in Section \ref{subsec:synt-test}.}

{\bluetext 
The average computation times for calculating point estimates and the corresponding confidence regions for the missing pixels were also evaluated. Using the (rescaled) \textit{Set12} dataset, we computed average results over $10$ different instances of missing pixels  for each image in the dataset, across varying percentages of observed data. Both the total computation time and the per-pixel time were measured. The averaged results are presented in Table \ref{tab-time-pe-uq}. It can be observed that increasing the number of missing pixels leads to faster computation of point estimates, since the dimension of the Gram matrix decreases. In contrast, while estimating confidence regions requires more computations as the number of missing pixels increases, the time of per-pixel computation is reduced due to the smaller dimension of the extended Gram matrix.}

{\renewcommand{\arraystretch}{1.6}
\begin{table}[!b]
{\bluetext
\caption{
Average computation times (in seconds) of the point estimate and the corresponding confidence region, as a function of the 
pixel removal percentage, for real-world inpaiting tasks.}
\begin{tabular}{|l||c|c|c|c|c|}
\hline
Pixel removal percentage  & $5\,\%$              & $10\,\%$            & $15\,\%$ & $20\,\%$ & $25\,\%$           \\ \hline
\hline
Full time (point estimate)   & $27.2587$  & $25.9469$ & $24.3020$ & $22.4343$ & $21.2460$\\ \hline
Per-pixel time (point estimate)  & $0.1331$ & $0.0633$ & $0.0396$ & $0.0274$ & $0.0207$ \\ \hline
Full time (confidence region) & $54.2986$ & $97.6210$ & $136.0579$ & $165.6048$ & $165.6631$\\ \hline
Per-pixel time (confidence region) & $0.2651$ & $0.2383$ & $0.2214$ & $0.2022$ & $0.1618$\\ \hline
\end{tabular}
\label{tab-time-pe-uq}
}
\end{table}
}

\begin{figure}[!t]
    \centering
	\hspace*{-2mm}		
	\includegraphics[width = \columnwidth]{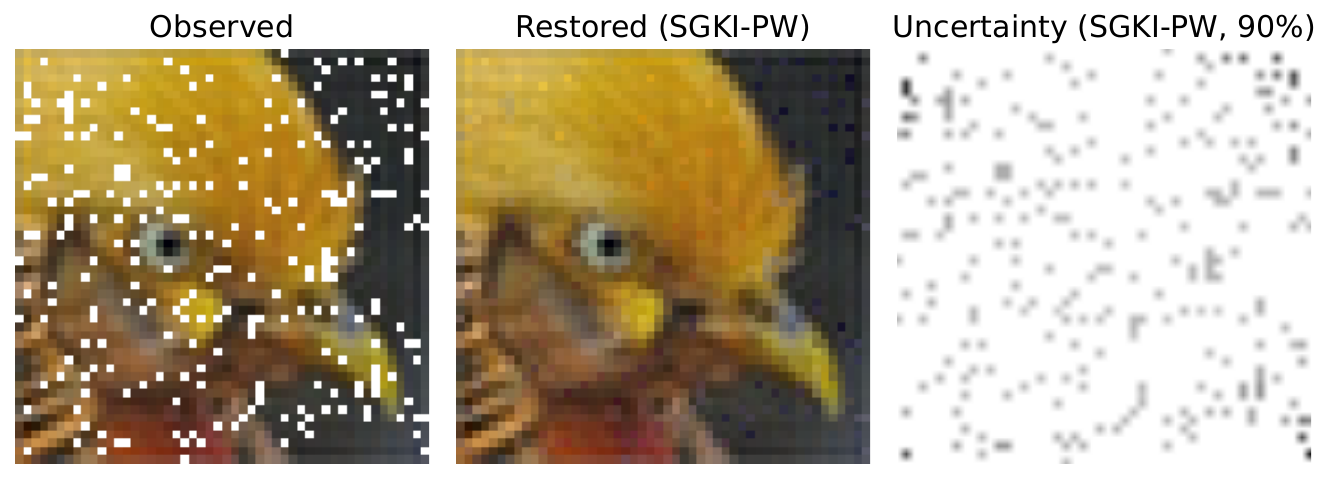} 	
    \caption{Single image inpainting and uncertainty quantification with $90 \%$ of the pixels observed.}
\label{fig:experiment6}
\end{figure}

Finally, our last experiment illustrates that SGKI can be applied for {\em color} images, as well, and it also illustrates the resulting confidence bands. Figure \ref{fig:experiment6} presents an experiment for {\em image inpainting} on a $50 \times 50$ {color} image together with the accompanied UQ regions. In this case $2250$ from the $2500$ pixels were observed, then the missing pixels were estimated with the minimum norm (PW) interpolant. SGKI was used with the Paley-Wiener kernel with parameters $\eta= 50$ and $\gamma = 0.1$ $($i.e., the confidence band had coverage probability $90\,\%)$. For a scalar output, the size of the uncertainty for query input $x_0$ can be defined as $|I_1(x_0) - I_2(x_0)|$. To help the visualization, a relative quantity was used: let us denote the largest uncertainty value by $U_{\max}$, then for an arbitrary $x_0 \neq x_k$, $k \in [n]$, the relative uncertainty was defined as $1-((|I_1(x_0) - I_2(x_0)| / U_{\max})^{1/4})$. These values were combined in the vector-valued (RGB) case by using the luminance weights: $0.3$ (R), $0.59$ (G) and $0.11$ (B). 

\section{Conclusions}
\label{sec:conlusions}

In this paper, we proposed a statistical learning approach to single image inpainting and super-resolution problems, extending the ideas of \citep{csaji2022nonparametric}. We call the method SGKI, as it is based on simultaneously guaranteed kernel interpolations. SGKI can construct simultaneous, non-asymptotic, non-parametric confidence bands under the core assumption that the underlying data-generating function is from a Reproducing Kernel Hilbert Space (RKHS) having a continuous and universal kernel. 

We showed that the uncertainty of the missing pixels can be quantified and a point estimate was also provided in the form of the  minimum norm interpolant. We proved that the SGKI interpolant is always included in the confidence band, we argued that the approach can be extended to vector-valued outputs, needed to handle color images, and a way to reduce the computational complexity of SGKI 
was also given, based on recursively computing the inverse of the kernel matrix using Schur complements. 

Several numerical experiments were also presented supporting the viability of the approach. For both the single image inpainting and super-resolution tasks, the SGKI method was compared with a variety of widespread methods. Quantitative results were given on a collection of synthetic images, which were randomly generated from a Paley-Wiener space, and also on real-world photos from the grayscale Set12 dataset. Finally, the uncertainty quantification capability of SGKI was illustrated and it was empirically demonstrated that SGKI can also be applied for color images.

Future research directions include generalizing the stochastic kernel norm bound construction of Section \ref{sec:kernelnorm} from Paley-Wiener kernels to other reproducing kernels and to extend the method to be able to efficiently handle noise reduction tasks, as well. {\blue It would also be beneficial to further investigate the reduction of the computational complexity of SGKI, possibly by allowing a trade-off between the computational requirements (speed) and the tightness of the achievable confidence bounds (precision).
More real-world datasets containing higher resolution images could also be tested.
}

\section*{Acknowledgements}
{\bluetext The authors thank Kriszti\'an Kis for his assistance in the numerical implementations.}
This research was supported by the European Union project RRF-2.3.1-21-2022-00004 within the framework
of the Artificial Intelligence National Laboratory, Hungary. The work was also partially supported by the TKP2021-NKTA-01 project of the National Research, Development and Innovation Office (NRDIO), Hungary.

\end{document}